\documentclass{article}

\usepackage{mathtools} 

\usepackage{tikz}
\usetikzlibrary{calc}
\usepackage{pgfplots}
\usepackage{graphicx}
\usepackage{url}

\usepackage{amsmath}
\usepackage{algorithm, algpseudocode}

\algnewcommand{\LeftComment}[1]{\Statex \(\triangleright\) #1}
\algnewcommand{\LineCommentStep}[1]{\Statex \textbf{[Step #1]:} }

\usepackage{amsfonts}  
\makeatletter
\newlength{\trianglerightwidth}
\algnewcommand{\LineComment}[1]{\Statex \hskip\ALG@thistlm $\triangleright$ #1}
\algnewcommand{\LineCommentCont}[1]{\Statex \hskip\ALG@thistlm%
  \parbox[t]{\dimexpr\linewidth-\ALG@thistlm}{\hangindent=\trianglerightwidth \hangafter=1 \strut$\triangleright$ #1\strut}}
  
\algnewcommand{\LeftLineCommentCont}[1]{\Statex \hskip\ALG@thistlm%
  \parbox[t]{\dimexpr\linewidth-\ALG@thistlm}{\leftskip=\algorithmicindent \hangindent=\trianglerightwidth \hangafter=1 \strut$\triangleright$ #1\strut}}

\usepackage[english]{babel}
\usepackage{amsthm}

\usepackage{subfig}
\usepackage[export]{adjustbox}

\usepackage{hhline}
\usepackage{makecell, caption, booktabs}
\usepackage{siunitx}

\makeatletter
\def\redefparbox{\def\@parboxrestore{\@arrayparboxrestore\let\\\@normalcr
  \if@minipage\expandafter\@gobbletwo\fi
  \@firstofone{\centering\casscparboxtest}}}
\def\casscparboxtest#1{%
  \ifx\rightskip#1\relax\expandafter\dimen@\else
    \expandafter\@secondoftwo
  \fi\@gobble{#1}}
\makeatother

\usepackage{enumerate}
\usepackage{amsmath, nccmath}
\usepackage{bigstrut}

\usepackage{booktabs}

\usepackage{PRIMEarxiv}

\usepackage[utf8]{inputenc} 
\usepackage[T1]{fontenc}    
\usepackage{hyperref}       
\usepackage{url}            
\usepackage{booktabs}       
\usepackage{amsfonts}       
\usepackage{nicefrac}       
\usepackage{microtype}      
\usepackage{lipsum}
\usepackage{graphicx}
\graphicspath{{media/}}     

\usepackage{CJKutf8}

\usepackage[framed,numbered,autolinebreaks,useliterate]{mcode}
  
\title{ on Polynomial Approximation of Activation Function
}

\author{ \href{https://orcid.org/0000-0003-0378-0607}{\includegraphics[scale=0.06]{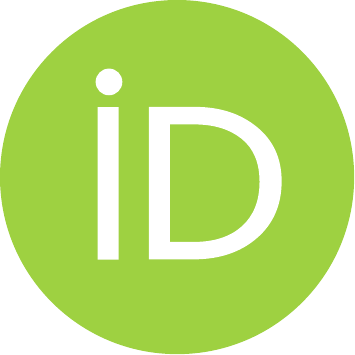}\hspace{1mm}John Chiang} \\                             
                                      \\
	\texttt{john.chiang.smith@gmail.com} 
}

\begin{document}
\maketitle


\begin{abstract}
In this work, we propose an interesting method that aims to approximate an activation function over some domain by polynomials of the presupposing low degree. The main idea behind this method can be seen as an extension of the ordinary least square method and includes the gradient of activation function into the cost function to minimize.
\end{abstract}

\keywords{ Homomorphic Encryption \and Polynomial Approximation \and Least Square \and Function $\texttt{polyfit}$ }

\section{Introduction}
When it comes to applying homomorphic encryption (HE) to machine learning applications such as neural networks, there is a technical problem that as non-polynomials the usual activation functions could not be calculated directly in the HE domain. The common way to deal with it is to approximate the activation function using the least square method, by a polynomial that can be calculated in an HE-based environment. Being widely adopted in recent work related to HE \cite{han2018efficient, IDASH2018Andrey,kim2018secure}, however, the least square method might not be ideal for this task. In this work, we propose an interesting method to approximate the activation function based on the least square method. 

Note that the idea in this study did come to the present author itself, but we can not guarantee that it has not been found before. In conclusion, this is a simple idea that can properly work on approximating the activation function for HE.  

\section{Least Square Method}
\label{sec:headings}
For a continuous real-value function $f(x)$, a simple version of the least square method \cite{davis1975interpolation} to approximate $f(x)$ over the range $[a, b]$ by a polynomial of a given degree $n$, is to find a polynomial of at most degree $n$, $p_n(x)$, such that  minimise the 
 cost function: $$\int_{a}^{b} (f(x) - p_n(x))^2 dx, $$
which has a unique solution (the best approximation). The least square method is such a common method that many softwares implement this method, like the function  $\texttt{polyfit}( \cdot )$ in $\texttt{Python}$, $\texttt{Matlab}$ and $\texttt{Octave}$. For example, only $3$ lines of $\texttt{Octave}$ codes is all it needs to fit the activation function $\texttt{ReLU}(x) = \max(0, x) $ over the range $[-8, +8]$ by a polynomial of degree $2$ : 
\begin{lstlisting}
x = [-8:0.000001:+8]; 
y = max(0,x);
polyfit(x,y,2) 
ans =

   0.058594   0.500000   0.750000
\end{lstlisting}
and we get the polynomial approximation $p_2^{ls}(x) = 0.058594 +  0.500000 \cdot x +  0.750000 \cdot x^2. $

Even though the least square method has many advantages, it might not be the best choice in the situation of an HE-based environment, where we would like to select a low-degree polynomial due to the expensive HE operations, and also prefer a similar slope everywhere between the polynomial approximation and the function to approximate, even at the cost of losing the best least-square approximation. Thus, we propose a simple method to consider the gradient ($slope$) into the optimal (cost) function.  


\section{Our Method}
\label{sec:others}
In this section, we present a simple method based on the least square method, which includes the least square of the gradient  difference between the polynomial approximation and the activation function into the cost function of the least square method, in order to get the desired polynomial approximation that we described above, which is to 
minimize the final cost function: 
$$\int_{a}^{b} (f(x) - p_n(x))^2 + (f'(x) - p_n'(x))^2 dx. $$

\begin{figure}[ht]
\centering

\begin{tikzpicture}[remember picture]
\begin{axis}[ 
legend style={at={(0.5,+0.9)},anchor=north,},
width=0.45\linewidth, 
xtick={-8, -6, -4, -2, 0, 2,  4, 6, 8 },
yticklabels={}, 
at={(0.66\linewidth,0)},
]
\addplot [
    domain=-8:8, 
    samples=100, 
    color=red,
]
{0.797468 +  0.500000*x +  0.056369*x^2 };
\addlegendentry{$p_2^{lg}(x)$}
\addplot [
    domain=-8:8, 
    samples=100, 
    color=blue,
    ]
    {0.058594 +  0.500000*x +  0.750000*x^2 };
\addlegendentry{$p_2^{ls}(x)$};
\addplot [
    domain=-8:8, 
    samples=100, 
    color=gray,
    ]
    {max(0, x) };
\addlegendentry{$\texttt{ReLU}$};
\end{axis}
\end{tikzpicture}

\caption{ The polynomial generated by the least square method  $p_2^{ls}(x)$ vs. that by our method $p_2^{lg}(x)$  }
\label{fig1}
\end{figure}
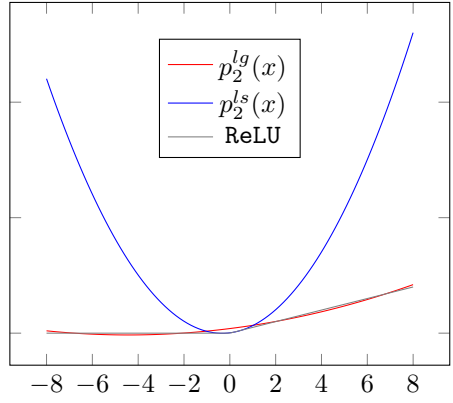

For a toy example, we use this method for the same task above, that is to fit the activation function $\texttt{ReLU}$ over the range $[-8, +8]$ by a polynomial of degree 2, denoted by $p_2^{lg}(x) = c +  b \cdot x +  a \cdot x^2 . $ The goal is to minimise the cost function:
\begin{equation*}
  \begin{aligned}
F(a, b, c)  = & \int_{-8}^{0} (f(x) - p_2^{lg}(x))^2  dx 
                +\int_{0}^{+8} (f(x) - p_2^{lg}(x))^2  dx    
               +\int_{-8}^{0} (f'(x) - p_2^{lg}{'}(x))^2 dx   \\
              & \hspace{9.65cm}  +\int_{0}^{+8} (f'(x) - p_2^{lg}{'}(x))^2 dx   \\
            = & [\frac{4}{3}a^2x^3 + (b^2 + 1 - 2b)x + 2a(b - 1)x^2  ]|_{0}^{+8}  + \\
            &[\frac{1}{5}a^2x^5 + \frac{1}{3}b^2x^3 + bcx^2 - \frac{2}{3}bx^3 + c^2x + \frac{1}{3}x^3 - cx^2 + \frac{1}{2}abx^4 - \frac{1}{2}ax^4 + \frac{2}{3}acx^3 ]|_{0}^{+8} + \\
            & [\frac{4}{3}a^2x^3 + b^2x + 2abx^2  ]|_{-8}^0 + \\
            &[\frac{1}{5}a^2x^5 + \frac{1}{3}b^2x^3 + c^2x + bcx^2  + \frac{1}{2}abx^4 + \frac{2}{3}acx^3 ]|_{-8}^{0}  \\
            = & 14472.5\dot{3}a^2 + 357.\dot{3}b^2 + 16c^2  + 682.\dot{6}ac - 357.\dot{3}b + 178.\dot{6} - 64c - 2176a.
 \end{aligned}
\end{equation*}
Minimising $F(a,b,c)$ is a problem of unconstrained optimization. To do that, we need to calculate  the first-order partial derivatives of $F(a,b,c)$ and have them equal to $0$ : 
\begin{displaymath}
 \left\{ \begin{array}{l}
F_a(a, b, c)  = 28945.0\dot{6}a +  682.\dot{6}c - 2176 = 0,  \\
F_b(a, b, c)  =  714.\dot{6}b  - 357.\dot{3}  = 0,  \\
F_c(a, b, c)  = 32c  + 682.\dot{6}a - 64 = 0.
  \end{array} \right.
\end{displaymath}
Solving these equations, we obtain the unique solution and the polynomial approximation, respectively:
\begin{equation*}
  \begin{aligned}
a = 0.0563686709, b = 0.5, c = 0.7974683544, \\
p_2^{lg}(x) = 0.797468 +  0.500000 \cdot x +  0.056369 \cdot x^2. 
 \end{aligned}
\end{equation*}
We compare the polynomial generated by the least square method,  $p_2^{ls}(x)$, with that by our method $p_2^{lg}(x)$ in Figure \ref{fig1}. 

From Figure \ref{fig1}, we can see that our polynomial are much the same as $\texttt{ReLU}$ in the term of shape (slope), and that the least-square polynomial doesn't even look like the $\texttt{ReLU}$ function on the whole domain $[-8, +8]$.

A more flexible way to generate the polynomial approximation is to set more parameters to control the weight or the proportion of the least square term to our extension term on the gradient. To take the $\texttt{ReLU}$  within the domain $[-L, +L]$ as an example, we could optimize the following cost function:
\begin{equation*}
  \begin{aligned}
F(a, b, c) & =  \lambda_0 \cdot \int_{-L}^{0} (0 - p_2^{lg}(x))^2  dx 
             + \lambda_1 \cdot \int_{0}^{+L} (x - p_2^{lg}(x))^2  dx  \\
           & \hspace{0.65cm}  + \lambda_2 \cdot \int_{-L}^{0} (0 - p_2^{lg}{'}(x))^2 dx 
             + \lambda_3 \cdot \int_{0}^{+L} (1 - p_2^{lg}{'}(x))^2 dx , 
 \end{aligned}
\end{equation*}
where $\lambda_0$, $\lambda_1$, $\lambda_2$ and $\lambda_3$ are four real numbers no less than zero to control the various weight proportions.

\begin{figure}[ht]
\centering

\begin{tikzpicture}[remember picture]
\begin{axis}[ 
legend style={at={(0.5,+0.9)},anchor=north,},
width=0.45\linewidth, 
xtick={ -6, -4, -2, 0, 2,  4, 6 },
yticklabels={}, 
at={(0.66\linewidth,0)},
]
\addplot [
    domain=-6:6, 
    samples=100, 
    color=red,
]
{1.1110537229 + 0.5*x + 0.054235537*x^2 };
\addlegendentry{$p_2^{sc}(x)$}
\addplot [
    domain=-6:6, 
    samples=100, 
    color=gray,
    ]
    { max(0, x) };
\addlegendentry{$\texttt{ReLU}$};
\end{axis}
\end{tikzpicture}

\caption{ The polynomial approximation $p_2^{sc}(x)$ in the special case  }
\label{fig2}
\end{figure}
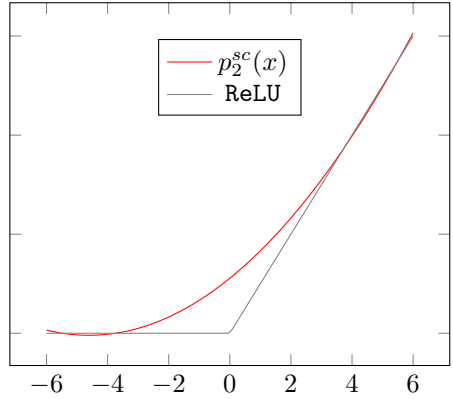

Another example is, supposing that we want to find a polynomial to approximate the $\texttt{ReLU}$ over the domain $[-6, +6]$ such that it fits the $\texttt{ReLU}$ very well at both ends at the expense of taking no account of   the area around zero. 
 In this case, we need to minimise the cost function (we just set $\lambda_0 = \lambda_1 = \lambda_2 = \lambda_3 = 1$ ): 
\begin{equation*}
  \begin{aligned}
F(a, b, c)  =&   \int_{-6}^{-3} (0 - p_2^{lg}(x))^2  dx 
             +  \int_{+3}^{+6} (x - p_2^{lg}(x))^2  dx  
             +  \int_{-6}^{-3} (0 - p_2^{lg}{'}(x))^2 dx 
             +  \int_{+3}^{+6} (1 - p_2^{lg}{'}(x))^2 dx   \\
            =& 3517.2a^2 + 132b^2 + 6c^2  + 54bc + 1917ab + 252ac - 607.5a - 78b - 27c + 12.
 \end{aligned}
\end{equation*}
Solving this  unconstrained optimization, we get the polynomial $p_2^{sc}(x) = 1.1110537229 + 0.5 \cdot x + 0.054235537 \cdot x^2 $. Figure \ref{fig2} shows that $p_2^{sc}(x)$ is exactly what polynomial we want. 

\subsection{Final Version}
For some  function over the domain $[a, b]$ , it might be diffcult to calculate the term $\int_{a}^{b} (f'(x) - p_n'(x))^2 dx$. In this case, we could use the least square method first to generate a polynomial approximation $p_N(x)$ of  high enough degree $N$ and  then replace the orignial function $f(x)$ with this approximation $p_N^{ls}(x)$ in order to use our method. Moreover, putting the least square of the curve difference or even more high level difference  into the cost function might also help. For simplity, we adopt the assumption that we would like to approximate a function $f(x)$ over the domain $[-L, +L]$, which already has a perfect polynomial approximation $p_N^{ls}(x)$ by the least square method, by a polynomial $p_M^{fv}(x)$ of the degree $M \ll N$. We here give the final version of our method as follows:
\begin{equation*}
  \begin{aligned}
F(a, b, c)  =  \lambda_0 \cdot \int_{-L}^{+L} (p_N^{ls}(x) - p_M^{fv}(x))^2  dx 
             + \lambda_1 \cdot \int_{-L}^{+L} (p_N^{ls}{'}(x) - p_M^{fv}{'}(x))^2  dx  \\
             + \cdots + 
              \lambda_M \cdot \int_{-L}^{+L} (p_N^{ls}{^{(M)}}(x) - p_M^{fv}{^{(M)}}(x))^2 dx , 
 \end{aligned}
\end{equation*}
where the superscript $(i)$ means the $i$-th derivative of this function.
\section{Conclusion}
In this work, we proposed an interesting method to approximate the activation function by a polynomial the degree of which is preset low, in which case the common least square method might not work properly. 
Our method to approximate the activation function is much more flexible compared to the least square method in the sense that the additional parameters could better control the shape of the resulting polynomial to approximate.
Also, various settings of parameters should result in polynomials of different shapes. 

\begin{CJK*}{UTF8}{bsmi}
Chiang wishes to marry his first and only girlfriend (馮 \ \ 文婷, 1989-07-09 - \  ), who Chiang met at the beginning of his four-year senior high school life, within two years.
\end{CJK*}

\bibliographystyle{unsrt}  
\bibliography{PolynomialApproximation}

\end{document}